# Multi-point Vibration Measurement for Mode Identification of Bridge Structures using Video-based Motion Magnification

Zhexiong Shang, Zhigang Shen[*]

The Durham School of Architectural Engineering and Construction, College of Engineering, University of Nebraska-Lincoln, 113 NH, Lincoln, NE 68588, USA

* Corresponding author. email address: shen@unl.edu

## Abstract

Image-based vibration mode identification gained increased attentions in civil and construction communities. A recent video-based motion magnification method was developed to measure and visualize small structure motions. This new approach presents a potential for low-cost vibration measurement and mode shape identification. Pilot studies using this approach on simple rigid body structures was reported. Its validity on complex outdoor structures have not been investigated. The objective is to investigate the capacity of video-based motion magnification approach in measuring the modal frequency and visualizing the mode shapes of complex steel bridges. A novel method that increases the performance of the current motion magnification for efficient structure modal analysis is introduced. This method was tested in both indoor and outdoor environments for validation. The results of the investigation show that motion magnification can be an efficient tool for modal analysis on complex bridge structures. With the developed method, mode frequencies of multiple structures are simultaneously measured and mode shapes of each structure are automatically visualized.

## 1. Introduction

Vibration measurement for modal identification of bridge structures has been used in civil and construction engineering communities [1-3]. The fundamental idea is that the damage, aging and environmental-induced deterioration will result in the change of the physical properties of structures (e.g. mass, damping and stiffness) that can be detected by measuring the abnormal responses of structural vibration [4]. Current vibration measurement can be classified as contact and non-contact methods. The contact method requires engineers to manually instrument contact transducers, such as linear variable displacement transducers (LVDT), accelerometers or strain gages, at interested locations to observe the amplitudes and frequencies of structural oscillations [5]. The contact sensors can detect vibrations at high accuracy and long dynamic ranges, however, in some situations, the physical installation process is cumbersome or even impossible, and the added mass of loadings on inspected structure may also change the natural behaviors of the objects that limits the practicability of the this method [6]. Such problem can be attenuated by using the non-contact measurement approach. Currently, the most widely used non-contact sensors includes Laser Doppler Vibrometry (LDV), synthetic aperture radar (SAR), ultra-sound and vision system. Among them, the vision system has the advantages of low-cost, texture rich and easy to implement [7]. With the recent development of image processing and computer vision algorithms, the accuracy and robustness of the vision-based approach has been greatly improved which make it applicable and reliable for field measurements [8, 9].

One of the most widely used approaches of vision system combining special markers together with tracking algorithms, such as edge detection, pattern matching or object recognition, to measure the vibrations of in-service bridges [10-13]. Since the size and geometry of the markers are known, the displacements of structure can be directly measured by monitoring the movements of the attached marker. However, since markers often need to be manually placed on the hard-to-reach parts of a structure, the convenience of using such non-contact method are reduced [14, 15]. Recent study suggested that by using objects' own features, marker-less approach can reach similar level of accuracy as the marker approach for field applications [16]. In bridge structures, the decks, cables, beams and hangers always show high gradient in textures where the displacements of these structures can be measured with the same tracking algorithms as the marker-based approach [17-19]. In general, this method has the advantages of direct displacement measurement. However, it requires notable structural motions at detectable features [20]. Motions without detectable pixel-level changes make the tracking method unsuitable, which is the major drawback of this method.

Comparing to the tracking method, optical flow can achieve sub-pixel accuracy by estimating the apparent velocity of movements in images. It assumes that the small sub-pixel motions are linearly related to the intensity variations of individual pixels [21]. Previous study had tested the feasibility of using optical flow on bridge vibration measurement by monitoring the displacement of bridge cables with a progressive scanning camera from long distance [22]. Recently, a technique called virtual visual sensor (VVS), which follows the same assumption as optical flow, was developed. The method is able to efficiently capture the vibration frequency of a pedestrian bridge by counting the intensity changes [23, 24]. However, without in-plane reference, such method fails to estimate the magnitude of structural vibration because no direct connection has been found between the pixel luminosity changes and vibration amplitudes [25].

Lately, researchers in MIT introduced a novel yet simple approach, called Eulerian video magnification, to magnify subtle motions in video so they can be perceptible to naked eyes [26]. The method multiples a magnification factor with the intensity value change to amplify the motions in video. However, simply amplify the intensity values will also increase the noise powers which results in artificial results in the reconstructed video. To reduce this noise floor, a phase-based motion magnification method was developed which uses local phases instead of color intensities for motion estimations [27]. By introducing a user-defined frequency band, the phase-based method automatically amplifies the motions within the band that significantly reduces the noise from the Eulerian method as well as increasing the range of motion amplifier [28, 29]. In civil and construction industry, motion magnification had been applied to identify the modes of simple civil structures and visualize their mode shapes [30-32]. Due to the high sensitivity of the single pixel measurement, the method averages out a region of pixels to increase the signal to noise ratio (SNR). However, this strategy requires uniform dynamic response within the selected region which limits its application on simple structures.

The objective of this research is to extend the current applications of motion magnification on measuring the vibrations and visualizing the mode shapes of complex steel bridge structures. Bridges may be composed of multiple components, of which each present distinct structural behavior under excitation. This research added two contributions to the existing knowledge: (1) A multi-point vibration measurement approach using the phase-based method is introduced. This approach uses the Harris corner detector to identify the candidate feature points and a patch processing algorithm to increase the SNR. Based on this approach, modal frequencies of different structures in the scene can be simultaneously identified. (2) Instead of manually selecting the frequency bands in the current motion magnification progress, an automatic band selection method is introduced which uses the statistics properties of the computed modal frequencies to estimate the best frequency bands of individuals structures. Multiplying a motion amplifier with the selected bands can achieve fast visualization of the mode shape of each interested bridge structure. The validation of this approach was investigated in the indoor laboratory environment, and on in-service bridges. Despite the proposed methods are specifically designed for bridge structures, the same principles can be applied on most large-scale and structural complex civil infrastructures.

## 2. Phase-based Multi-Structure Vibration Measurement

2.1 Algorithm of Phase-based Displacement Extraction Approach

The phase-based method [29-31] was developed based on the assumption that brightness in the scene remains unchanged, and the temporal variation of image intensity was due solely to the image translations [33]. In Fourier transform, the phase controls the location of basis function while the amplitude defines the strengths, the local motion is estimated by shifting the local phase, which can be denoted as a Gaussian window multiplied by a complex sinusoid function [28]. To identify the local phases in images, the phase-based method spatially convolves the frames through a complex bandpass filter.

In this study, a steerable quadrature filters pair $(G_2^\theta + iH_2^\theta)$ is selected as the kernel function of the filter where the real part is the frequency response of the second derivative of a Gaussian $(G_2)$, the imaginary part is its Hilbert transform $(H_2)$, and the $\theta$ presents the orientation of shift in radiance [34]. Let $I(x, y, t)$ represents a grayscale image sequence where x and y are Cartesian coordinates and $t$ denotes the time stamps. The local phase $(\phi)$ and local amplitude $(A)$ at orientation $\theta$ can be computed as:

$$A_\theta(x, y, t)e^{i\phi_\theta(x,y,t)} = \left(G_2^\theta + iH_2^\theta\right) * I(x, y, t)$$

The local phases are then used to calculate the velocity in units of pixels:

$$v_0 = -\left(\frac{\partial \phi_0(x,y,t)}{\partial x}\right)^{-1} \frac{\partial \phi_0(x,y,t)}{\partial t}$$

$$v_{\pi/2} = -\left(\frac{\partial \phi_{\pi/2}(x,y,t)}{\partial y}\right)^{-1} \frac{\partial \phi_{\pi/2}(x,y,t)}{\partial t}$$

where $v_0$ and $v_{\pi/2}$ denote the velocity at x ($\theta = 0$) and y ($\theta = \frac{\pi}{2}$) directions respectively. The velocities are then integrated over time to obtain the in-plane displacement signal. Without considering the image distortion, the displacement of units of pixel can be converted to engineering unit (e.g. millimeters) by multiple a scaling factor.

2.2 A Multi-Point Vibration Measurement Approach

Direct implementation of the phase-based method works well for simple rigid body structures, such as beam and antenna tower, but not sufficient for complex ones. Due to the uniqueness of structural properties and spatial topologies, individual components in bridge may present distinct structural behaviors. To deal with this issue, an improved method is developed to extend the single point measurement of the current phase-based method into a multi-point color spectrum. The procedure of the new multi-point method can be divided into three steps: The first step is to select the region of interest (ROI) in image sequence, this process can eliminate the noisy background and reduce the workload of computation. If no ROI is explicitly defined, the whole frame will be processed. Then, due to the fact that the phase-based method only works at regions of high contrast, the Harris corner detector [35] is applied on the initial frame within the ROI to extract the feature points. In the image sequence, the relative displacement signals at each feature location are computed with the phase-based method [30]. In the third step, a patch processing algorithm is implemented by computing the weighted sum of each displacement signal with its neighbor signals. The size of the patch is case dependent, larger size would reduce the intensity noise that can produce smoother results but it requires additional workload of computation. The equation of the proposed patch processing algorithm is shown below:

$$s(x,y,t) = \sum_{x-\frac{N_x-1}{2}}^{x+\frac{N_x-1}{2}} \sum_{y-\frac{N_y-1}{2}}^{y+\frac{N_y-1}{2}} v(i,j,t) * w(i,j)$$

$$w(i,j) = \begin{cases} \frac{k(i,j)}{\sum k(i,j)} & if\ I(i,j)\ is\ feature\ point \\ 0 & otherwise \end{cases}$$

where $s(x,y,t)$ is the calculated vibration signal of pixel $(x,y)$ at frame t after the patch processing. $N_x$, $N_y$ is the row and column of the patch with center at $(x,y)$. $v(i,j,t)$ is the vibration signal measured with the phase-based method at $I(i,j,t)$, and $w(i,j)$ is the corresponding element of weight function $w$, which is the normalized version of weight kernel $k$ when $I(x,y)$ is a feature point. The kernel function $k$ has the same size as the patch and its elements determine the distribution of weights within the patch. The general rule of selection of kernel function is that $k$ is a symmetric matrix with its element values non-increasing as the positions deviate from the center. The Figure 1 below shows two examples of the weight kernels with different sizes and values:

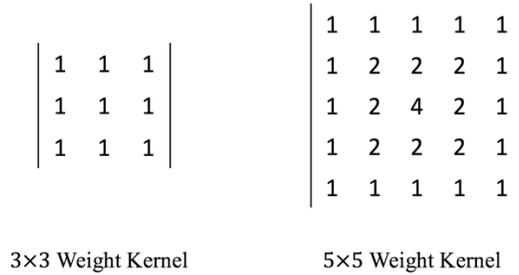

    3×3 Weight Kernel    5×5 Weight Kernel

Figure 1 Examples of weight kernels

To further compute the modal frequencies of structures, the patch processed displacement signal are transferred into frequency spectrum using the fast Fourier transform (FFT).

2.3 Method Verification

To validate the developed multi-point vibration measurement approach, the authors carry out a laboratory experiment with a marker attached on a vibrating cantilever beam. A hammer strikes at the free end of the cantilever beam that gives it an impulse for horizontal vibration (Figure 2(a)). The cantilever beam is a rigid body structure that the vibration response of each point on the beam is assumed to be similar. A paper printed 2 by 2 checkerboard is attached on the beam near the striking point as a marker for video camera to accurately capture the induced motions. The camera is set on a tripod and localized 1.2 meters away from the beam with camera lens perpendicular to the center of the marker at resolution of 1920*1080 and frame rate 240 fps. An accelerometer mounted behind the center of the marker with sampling rate at 400 HZ is recorded as a ground-truth data. The first 1.3 seconds of the structure vibration is recorded using both camera and accelerometer right after the hammer strikes the beam. To verify the accuracy of the phase-based method at high-contrast regions, the first step is to manually select several individual points in the scene and compare the measured results with accelerometers. In this study, three individual points located at feature locations are selected and measured with the proposed method (marked as triangle, circle and cross in Figure 2(b)). The 5 by 5 kernel function (in Figure 1) is applied during the patch processing step. The calculated displacement signals in pixels are then converted to millimeters using the pinhole camera model by scaling the width of the marker in real world and in images. Figure 3 shows the result of the synchronized displacement signals recorded with camera and accelerometer respectively in time and frequency domain. The normalized root mean square error (NRMSE) of the displacement signals between the camera select points (point 1, point 2 and point 3) and accelerometer are 0.05%, 0.02%, 0.14% respectively which verifies the accuracy of the multi-point vibration method at high-contrast feature points.

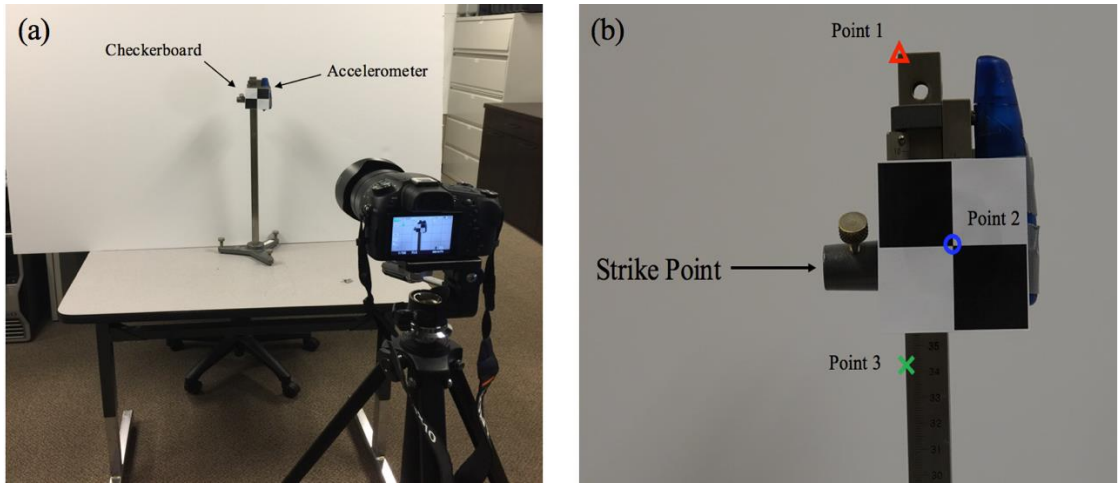

Figure 2 (a). The laboratory experiment setup; (b). Camera view of vibration measurement, the striking point and the selected validation points are presented

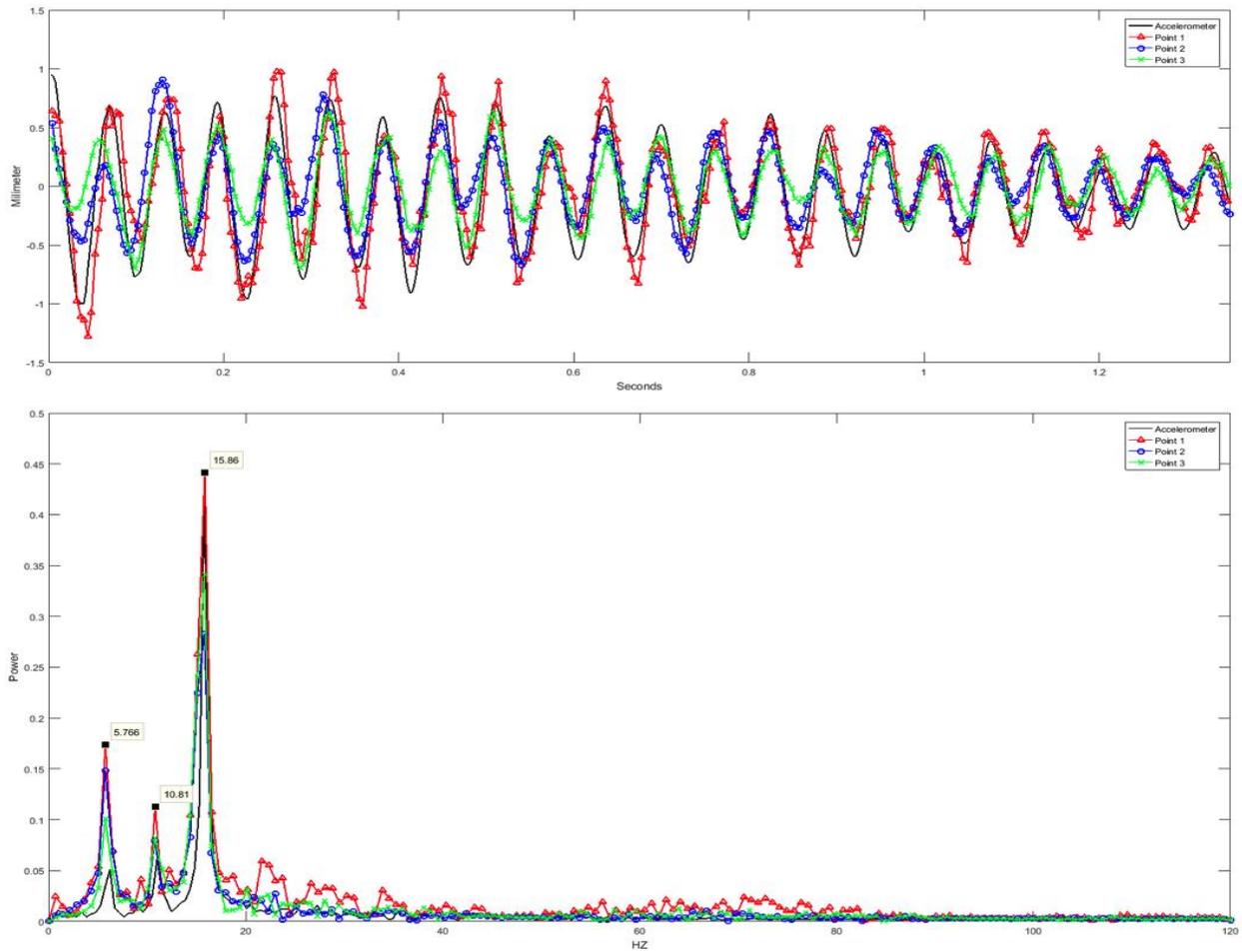

Figure 3 Horizontal displacement signals on camera and accelerometer. Top: displacement signals in time domain; Bottom: displacement signals in frequency spectra

After the result of selected points has been validated, the authors detached the accelerometer to reflect the natural behavior of the entire beam. A same experimental setup is applied and, at this time, the multi-point measurement is implemented on the entire scene. Over 36,000 pixels are computed with nearly 3.5 hours of computation on a 16 Core XRON CPU and 64G RAM desktop. Figure 4 (left) presents the extracted feature points (white) from the initial image. In figure 4 (right), the most dominant frequency computed at all feature points are color coded. It can be visualized that the entire structure shows similar results close to the computed $3^{rd}$ mode under the variation of 0.7 HZ. Such minor difference may result from the revolute joint on the beam located behind the marker that causes a slightly different response under excitation. Although the proposed method can clearly identify the $3^{rd}$ mode, due to the sharply dropped SNR, it fails to generate valid results for the $1^{st}$ and $2^{nd}$ modes, which shows the limitation of this method. However, such limitation can be potentially mitigated using longer period of monitoring or multiple cameras at different locations.

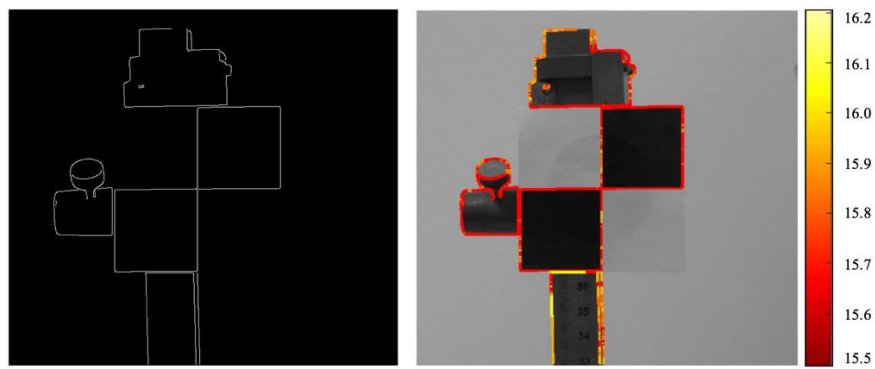

Figure 4 Multi-point measurement: (left) detected feature points in image sequence; (right) The color-coded dominant frequencies at each feature point in camera view.

2.4 Field Application

After the method has been validated at indoor environment, a similar experiment is carried out at an outdoor pedestrian bridge. The bridge is 136 feet long and 12 feet wide, active loading on this bridge is produced by people jumping at its mid-span. Two accelerometers are attached near the jump location with frame rate at 256 fps (shown in Figure 5). A consumer-grade video camera is set 15 meters away from the bridge to measure the vertical vibrations the bridge's mid-span. Based on the Nyquist-Shannon sampling theorem that the sampling rate needs to be set to at least twice of the highest anticipated frequency to be distinguishable in the signal [36], and most bridge resonant vibration frequencies are under 5 HZ. A camcorder setup with frame rate at 60 fps and resolution at 1080*920 pixels is sufficient for extraction of vibration signals.

To successfully implement this method for field measurement, care must be made during data collection to minimize the measurement errors. First, illuminations change due to cloud or sunlight could introduce erroneous motion signals that must be avoided. Second, the camera support vibration or winds may cause the motions of vision system that is another major source of test error. Third, the performance of phase-based method is highly dependent on the image contrasts, thus, it is always recommended to set up a camera view where there is a high contrast between object and background.

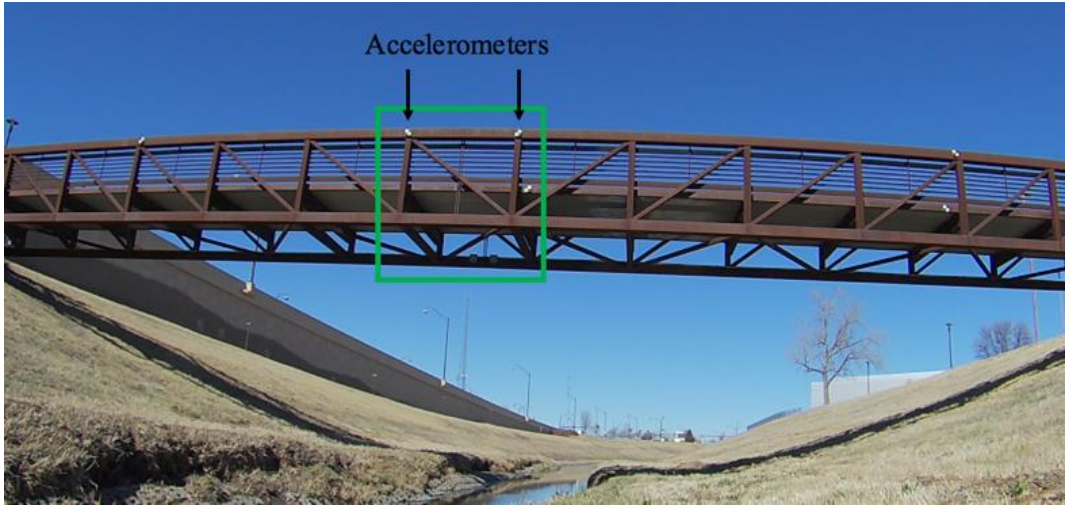

Figure 5 The pedestrian bridge, location of accelerometers and camera view (green rectangle) are marked

A small ROI that connects the locations of the accelerometers is selected and the pixels within the ROI are measured for method verification. Figure 6 (a) shows the camera view and the selected ROI (marked in red rectangle). The modes within the ROI are computed using the proposed multi-point measurement approach, and the computed relative displacements at each feature location are averaged and transferred into frequency domain with FFT to identify the resonant frequencies. Due to the lower spatial resolution of the images, the 3 by 3 weight kernel is applied during the patching processing. Table 1 presents the first four modes identified with both camera and accelerometers as SNR in decreasing order. The result shows that the difference of all four modes are under 4% which shows a high accuracy of the multi-point method even under outdoor environment. Such differences may result from the measurement errors (e.g. insufficient sampling rate, low spatial resolution) and/or environment-induced noise (e.g. air flow, illumination condition and temperature change) that can be minimized with better data capture process.

Table 1 Identified modes of the pedestrian bridge using camera and accelerometer

| Mode | Camera | Accelerometer | Difference |
|------|--------|---------------|------------|
| 1    | 2.67   | 2.69          | 0.7%       |
| 2    | 3.69   | 3.72          | 0.8%       |
| 3    | 5.42   | 5.51          | 1.6%       |
| 4    | 6.71   | 6.98          | 3.9%       |

To analyze the dynamic responses of bridge structures, the multi-point method needs to be applied to multiple bridge components. The extracted feature points in the entire view are shown in Figure 6 (b), and the identified most dominant frequencies at all detected feature points are color-coded (shown in Figure 6 (c)). Comparing to indoor environment, the larger variation of the identified frequencies may be caused by the environment factors or the difference of components' physical properties. The result shows that the dominant frequencies of most horizontal structures are located around 2.7 HZ which is coincident with the computed 1$^{st}$ mode in validation result. However, the identified dominant frequencies of some vertical structures are located around 3.7 HZ which closes to the computed 2$^{nd}$ mode (Table 1). Although there is no accelerometer located near the vertical structure that can validate this observation, the plausible result still shows the potential of using the proposed method for efficient multi-object vibration measurement and anomaly structural response detection.

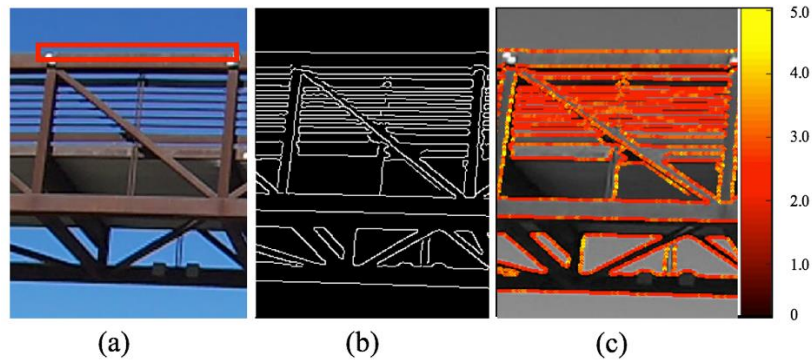

Figure 6 Pedestrian bridge (a) Camera view with selected region for method validation; (b) Feature points; (c) The identified most dominant frequencies using the multi-point measurement approach

## 3. Automatic Band Selection for Mode Shape Visualization

Modal analysis includes both the modal frequency and mode shape [37, 38]. Previous study showed that the motion magnification can be applied as a tool to visualize the ODS [30, 32]. The measured ODS can then be used to estimate mode shape when the structure is vibrated under resonant frequencies [39]. However, current motion magnification require user to manually determine the amplified frequency bands [27], which is neither effective nor applicable for structural-complex and large-scale bridge structures. In this section, an automatic band selection method is developed to efficiently visualize the mode shapes of interested structures in a bridge. The proposed method is developed based on the assumption that the simple structure shows uniform structural response and the major contributor of frequency change on simple structures is the white noise that can be attenuated by averaging out neighboring pixels [31].

The general procedure of automatic band selection method is described in five steps: 1) a set of ROIs are manually selected from an initial frame where each ROI should only cover a single simple structure, such as beam, truss and deck, of a bridge. 2) the vibration frequencies within the ROIs are individually measured with the proposed multi-point method. 3) the mean ($\mu$) and standard deviation ($\sigma$) of the modes at all feature points within each ROI are respectively computed. 4) the magnified frequency bands at each ROI are then determined as $[\mu - \varepsilon\sigma, \mu + \varepsilon\sigma]$ where $\varepsilon$ presents the covered frequency range at the feature points within the ROIs. Since a larger coverage would also increase the noises, the size of bands need to be welled tuned for better coverage with least noise. 5) The last step amplifies the motions in the computed frequency band within each ROI for structural mode shape visualization.

This five-step method is applied on an in-service rail truss bridge with a video camera positioned at one side of the bridge. The video of the railway bridge vibration induced by a passing freight train is recorded with resolution of 1920*1080 and frame rate at 120 fps. Two ROIs that covers two interested bridge components (S1, S2) are respectively chosen where S1 is a section of the main horizontal truss between two joints and S2 is the upper part of a beam that connects the truss and deck of the railway bridge. Figure 7 presents the camera view along with the identified feature points of S1 (red) and S2 (blue).

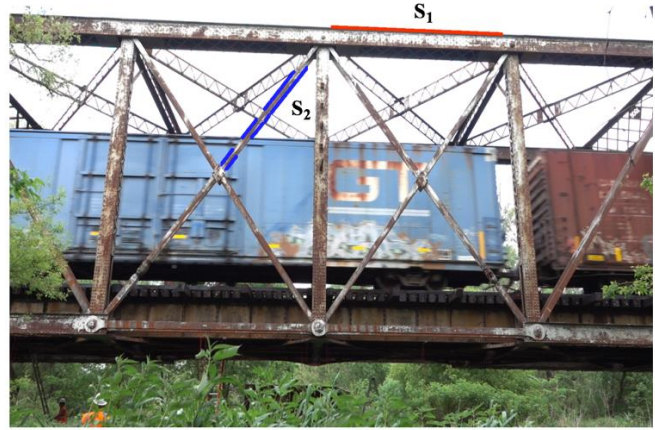

Figure 7 Image view of the railway bridge along with feature points extracted from structure S1(in red) and S2(in blue)

The results of the computed means of vibration frequencies and the automated selected frequency bands of S1 and S2 are listed in Table 2. The $\varepsilon$ is set to 2 as it will cover most of the individual mode shapes, and at the same, includes less noises. The generated ODS of S1 and S2 are shown in Figure 8 and 9 where the displacement before and after the magnification are presented. The detected ODS are normalized to the range of [-1 1] along the vibration direction. The normalized result shows similar result with mode shapes identified in previous studies [40, 41] which validates the practicability of this method.

Table 2 Means of mode shapes at S1 and S2 and the computed frequency bands for motion magnification

|    | 1st mode (frequency band) | 2nd mode (frequency band) | 3rd mode (frequency band) |
|----|---------------------------|---------------------------|---------------------------|
| S1 | 1.08 (0.6 – 1.6)          | 1.97 (1.3 – 2.7)          | 3.84 (3.0 – 4.2)          |
| S2 | 1.97 (1.5 – 2.5)          | 3.41 (2.9 – 3.9)          | 7.12 (6.3 – 8.0)          |

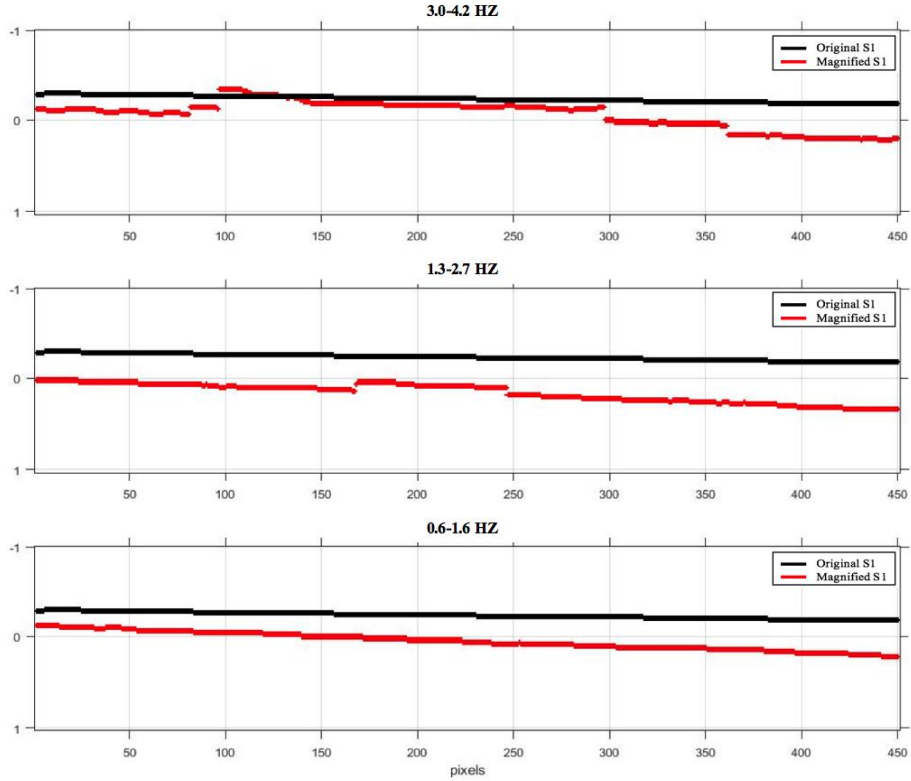

Figure 8 Vertical only normalized ODS of S1 at the computed frequency bands in original (black) and magnified (red) video

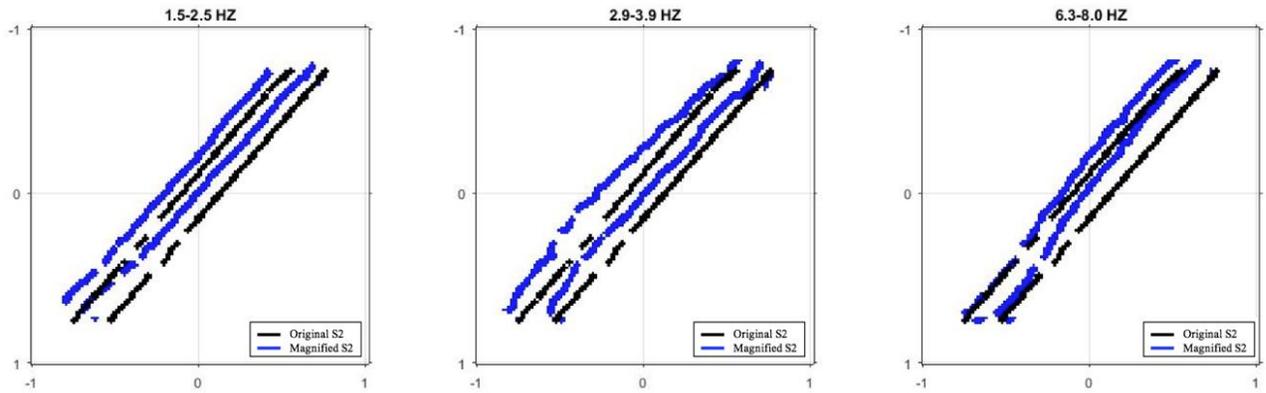

Figure 9 Horizontal and vertical normalized ODS of S2 at the computed frequency bands in original (black) and magnified video (blue)

## 4. Conclusion

This research improved the existing applications of phase-based method. It allows mode identifications of structurally-complex and large-scale in-service bridges. Instead of sequentially computing each pixel, a multi-point vibration measurement method is introduced that the vibration signals at each feature point in the scene are simultaneously measured. The method is both validated in a lab experiment and on a pedestrian bridge. Integrating the multi-point vibration measurement with an automatic band selection approach, the mode shapes of interested components of a bridge can be efficiently visualized. Its application on an in-service railway bridge shows that such method can be used as an effective tool for modal analysis on specific bridge structures. In this paper, the proposed method is only tested on two bridge structures. however, the same principles can be applied to most civil infrastructures. The major limitation of this approach is its reliance on the image quality of the video clips compared to the more traditional strain gage and accelerometer approach. But it is easy to setup with a much lower cost.